\documentclass[sigconf]{acmart}
\renewcommand\footnotetextcopyrightpermission[1]{} % removes footnote with conference information in first column
\usepackage{amsmath}
\usepackage{booktabs}
\usepackage{algorithm}
\usepackage{algorithmic}
\usepackage{makecell}
\usepackage{siunitx, mhchem}
\usepackage{multirow}
\usepackage{amsthm}
\usepackage{graphicx}
\usepackage{hyperref}
\graphicspath{ {graphs/} }
\usepackage{longtable}
\usepackage{tcolorbox}

\theoremstyle{definition}
\newtheorem{definition}{Definition}[section]

\AtBeginDocument{%
  \providecommand\BibTeX{{%
    \normalfont B\kern-0.5em{\scshape i\kern-0.25em b}\kern-0.8em\TeX}}}

\begin{document}

\title{Information-theoretic Evolution of Model Agnostic Global Explanations}


\author{Sukriti Verma}
\authornotemark[1]
\affiliation{%
  \institution{Adobe Systems}
  \city{Noida}
  \state{Uttar Pradesh}
 \country{India}
}
\email{sukrverm@adobe.com}

\author{Nikaash Puri}
\authornotemark[1]
\affiliation{%
  \institution{Adobe Systems}
  \city{Noida}
  \state{Uttar Pradesh}
 \country{India}
}
\email{nikpuri@adobe.com}

\author{Piyush Gupta}
\authornote{All authors contributed equally to the paper.}
\affiliation{%
  \institution{Adobe Systems}
  \city{Noida}
  \state{Uttar Pradesh}
 \country{India}
}
\email{piygupta@adobe.com}

\author{Balaji Krishnamurthy}
\authornotemark[1]
\affiliation{%
  \institution{Adobe Systems}
  \city{Noida}
  \state{Uttar Pradesh}
 \country{India}
}
\email{kbalaji@adobe.com}



\begin{abstract}
Explaining the behavior of black box machine learning models through human interpretable rules is an important research area. Recent work has focused on explaining model behavior locally i.e. for specific predictions as well as globally across the fields of vision, natural language, reinforcement learning and data science. We present a novel model-agnostic approach that derives rules to globally explain the behavior of classification models trained on numerical and/or categorical data. Our approach builds on top of existing local model explanation methods to extract conditions important for explaining model behavior for specific instances followed by an evolutionary algorithm that optimizes an information theory based fitness function to construct rules that explain global model behavior. We show how our approach outperforms existing approaches on a variety of datasets. Further, we introduce a parameter to evaluate the quality of interpretation under the scenario of distributional shift. This parameter evaluates how well the interpretation can predict model behavior for previously unseen data distributions. We show how existing approaches for interpreting models globally lack distributional robustness. Finally, we show how the quality of the interpretation can be improved under the scenario of distributional shift by adding out of distribution samples to the dataset used to learn the interpretation and thereby, increase robustness. All of the datasets used in our paper are open and publicly available. Our approach has been deployed in a leading digital marketing suite of products. 
\end{abstract}
%often capture patterns of the training data distribution that do not hold on test data distributions
%Change this############
% <concept_id>10010147.10010178</concept_id>
% <concept_desc>Computing methodologies~Artificial intelligence</concept_desc>
% <concept_id>10010147.10010257</concept_id>
% <concept_desc>Computing methodologies~Machine learning</concept_desc>
% <concept_id>10010147.10010341.10010342.10010344</concept_id>
% <concept_desc>Computing methodologies~Model verification and validation</concept_desc>

% \ccsdesc[500]{Computing methodologies~Artificial intelligence}
% \ccsdesc[500]{Computing methodologies~Machine learning}
% \ccsdesc[500]{Computing methodologies~Model verification and validation}

% \keywords{Model Interpretation, Interpretable Machine Learning}

\maketitle
\pagestyle{plain}

\section{Introduction}\label{introduction}

Complex machine learning models have been shown to be highly accurate and desirable towards many applications, from health to digital marketing. It is becoming increasingly important that experts be enabled to understand and explore the behavior of these models in a human interpretable way \cite{lipton2016mythos, ribeiro2016model, doshi2017towards}. However, the proprietary nature and the complexity of these models makes this a challenging problem to solve \cite{ribeiro2016should}. Hence, the field of interpretable machine learning has seen a resurgence in recent years. One area of the field focuses on creating models that are inherently interpretable to begin with \cite{angelino2017learning, letham2015interpretable, wang2015falling}. Other areas focus on post hoc interpretation of black box models. These methods can be classified into three categories:
\begin{itemize}
    \item Model Dependent: Exploiting model specific characteristics for interpretation \cite{foerster2017input, selvaraju2016grad, goyal2016towards, sundararajan2017axiomatic, setiono2004approach, shrikumar2016not}
    \item Local Explanations: Explaining individual model predictions \cite{ribeiro2016should, lundberg2017unified, ribeiro2018anchors, NIPS2018_7518, melis2018towards, koh2017understanding}
    \item Global Explanations: Explaining high-level global model behavior \cite{lakkaraju2016interpretable, lakkaraju2019faithful, bastani2017interpreting} 
\end{itemize}

\begin{table}[]
\centering
\caption{Sample interpretation generated by our approach, MAGIX, for a Random Forest model trained on the Mushroom Dataset \cite{Dua:2017}}
\begin{tabular}{|p{.27\textwidth}|p{.17\textwidth}|}
\hline
\textbf{Rule Clause} & \textbf{Model Prediction} \\ \hline
IF stalk-surface-above-ring = silky AND gill-spacing = close & THEN class = poisonous                \\ \hline
IF stalk-surface-above-ring = smooth                         & THEN class = edible                   \\ \hline
IF odor = foul AND ring-number = one                         & THEN class = poisonous                \\ \hline
IF odor = none                                               & THEN class = edible                   \\ \hline
\end{tabular}
\label{table:sample-rules-MAGIX}
\end{table}

While understanding individual predictions made by the model is useful, it is also important for decision makers to understand global patterns that the model uses for predictions. These patterns may be described using raw features or interpretable features \cite{conf/icml/KimWGCWVS18, NIPS2018_8003}. For instance, consider the global interpretation shown in Table \ref{table:sample-rules-MAGIX} for a model trained on the Mushroom dataset \cite{Dua:2017}. The rules generated by our approach provide a view into the high level patterns used by the model for decision making. For example, the rule `If odor = none then predict edible' shows that whenever the mushroom has no odor, the model classifies it as being edible. Such patterns could be useful to evaluate a black box model prior to deployment. Further, these patterns might reveal interesting patterns present in the original dataset.    

To evaluate whether these rules represent patterns used by the model, we use them to make predictions for previously unseen data points and measure the fraction of instances for which the model prediction matches the prediction made by the rules (Section \ref{results}). We describe an approach called \textbf{MAGIX} or \textbf{M}odel \textbf{A}gnostic \textbf{G}lobally \textbf{I}nterpretable E\textbf{x}planations to interpret black box machine learning classification models as human understandable rules. We share the same objective as existing approaches like MUSE \cite{lakkaraju2019faithful} that also explain the high-level global behavior of any given black box. However, we propose a mutual information based rule quality measure and optimize it using an evolutionary algorithm. We show in Section \ref{results} that MAGIX outperforms existing approaches across several datasets.   

\begin{table}[]
\centering
\caption{Sample dataset to illustrate distributional shift and the importance of uncertainty analysis}
\begin{tabular}{|c|c|c|c|}
\hline
\textbf{S. No.} & \textbf{Age} & \textbf{State} & \textbf{Model-Prediction} \\ \hline
1              & 27           & California     & not-default               \\ \hline
2              & 22           & Texas          & default                   \\ \hline
3              & 31           & California     & not-default               \\ \hline
4              & 21           & Texas          & default                   \\ \hline
\end{tabular}
\label{table:introduction-toy-dataset}
\end{table}

Recent work has demonstrated post hoc model interpretation to be unstable under the scenario of distributional shift \cite{ghorbani2019interpretation, lakkaraju2020fool}. In Section \ref{Uncertainty Analysis} we introduce a new parameter to evaluate the distributional robustness of a given interpretation. It is a measure of how well the interpretation is at predicting model behavior for previously unseen data distributions. Its importance can be understood intuitively by considering the sample dataset in Table \ref{table:introduction-toy-dataset}. In this case, a global explanation of model behavior could contain the rule `\textit{If Age $>$ 25 then the model predicts not-default}'. However, say the model was using the decision rule `\textit{If State = California then predict not-default}'. In that case, the rule `\textit{If Age $>$ 25 then the model predicts not-default}' also correctly explains model behavior. This is because age and state are correlated in the dataset. Measuring how well the interpretation predicts model behavior on a sample of the dataset may not uncover such errors. This uncertainty problem occurs due to the existence of multiple explanations for model behavior on the original data distribution, with each explanation having the same fidelity \cite{lakkaraju2020fool}. It has been discussed in the context of local explanations \cite{zhang2019should, ghorbani2019interpretation}. We introduce this in the context of global explanations by devising an evaluation strategy that measures model imitation on perturbed versions of the original dataset. We show how the quality of the interpretation falls when imitation is measured on data distributions that the model has not seen previously (Section \ref{Uncertainty Analysis}). Further, we show how the robustness of the interpretation to such perturbations can be improved by adding synthetic out of distribution samples to the dataset used to learn the interpretation (Section \ref{Ablation Study 3}).

Our contributions are as follows:
\begin{enumerate}
    \item We propose an approach, MAGIX, that explains the high-level global behavior of any black box classification model trained on numerical and/or categorical data. The interpretation produced by MAGIX consists of human understandable rules. These rules are evolved using a genetic algorithm guided by a novel \textbf{mutual information based fitness measure} (Section \ref{mi}). We show how our approach outperforms existing approaches on ten different publicly available datasets (Section \ref{results}).
    \item We introduce a new parameter to evaluate distributional robustness \cite{ghorbani2019interpretation, lakkaraju2020fool} of existing approaches. We show how existing approaches lack robustness when applied to previously unseen data distributions (Section \ref{Uncertainty Analysis}). Further, we show how the distributional robustness can be improved by adding synthetic out of distribution samples to the dataset used to learn the interpretation (Section \ref{Ablation Study 3}).
\end{enumerate}

\begin{figure}
\includegraphics[width=0.45\textwidth]{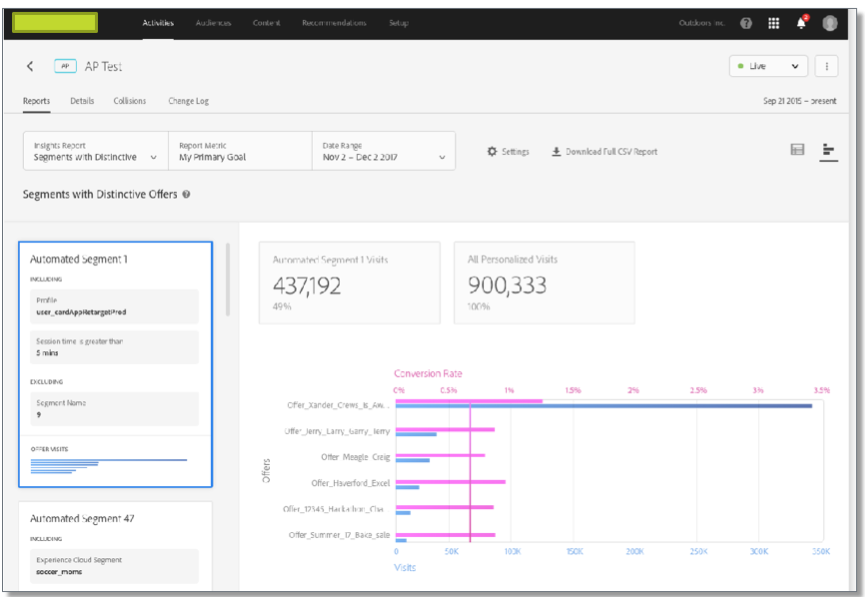}
\centering
\caption{MAGIX deployed in a leading Digital Marketing Platform as Insights Report helps Marketers to understand the high level strategies being used by the previously black box personalization offering, increasing adoption and trust.}
\label{figure:digital-marketing-insights-report}
\end{figure}

Our approach, MAGIX, has been deployed in a leading digital marketing platform in the form of an `Insights Report' (Section \ref{case-study-magix-insights-report}). Marketers use the digital marketing platform to personalise content at the user level by learning from past user behavior. Marketers were hesitant to use the black box personalization offering without understanding the high level strategies (global rules) that the model used for decision making. Therefore, MAGIX has been deployed in the platform to solve this problem. The machine learning model runs daily, followed by MAGIX, which generates an `Insights Report' that contains rules that explain model behavior. This report can be accessed by the marketer from the platform user interface (Figure \ref{figure:digital-marketing-insights-report}). The model is trained on user profile attributes. Therefore, each rule represents a user segment that the model learned for serving content. Each user segment is represented by a card on the left (`Automated Segment 1', `Automated Segment 47'). For each segment, the interface shows the offers shown to users from that segment (blue bar charts) along with corresponding click through rates (pink bar charts). For instance, most users in `Automated Segment 1' were served the offer 'Offer-Xander' by the model. These user segments are used by marketers as described in Section \ref{case-study-magix-insights-report}.

\section{Related Work}\label{lit_survey}

One set of approaches towards model interpretation take the route of learning predictive models which are human understandable to begin with such as linear
models, decision trees, decision lists/sets and generalized additive models \cite{letham2015interpretable, wang2015falling, lakkaraju2016interpretable, angelino2017learning, kim2019learning}. Other approaches focus on learning post hoc interpretations of black box models. These approaches can be classified across two dimensions:
\begin{itemize}
    \item Model Agnostic v/s Model Dependent
    \item Local Explanations v/s Global Explanations
\end{itemize}

There are various approaches that are model specific, that is they exploit model specific properties to construct explanations \cite{setiono2004approach, selvaraju2016grad, goyal2016towards, shrikumar2016not, sundararajan2017axiomatic,  foerster2017input,}. These are orthogonal to our work since our objective is to explain a black box model without exploiting its internal modelling characteristics i.e. a model agnostic approach. 

There are several approaches that explain model behavior locally (in a limited region of the input space) \cite{baehrens2010explain, lundberg2016unexpected, shrikumar2016not, lundberg2017unified, NIPS2018_7518}. LIME \cite{ribeiro2016should} explains the model prediction for one particular instance. In Anchors \cite{ribeiro2018anchors}, the authors show that explanations in the form of rules are more interpretable for humans than those produced by LIME. Local approaches such as LIME and Anchors derive rules that correctly explain a small region of the input space with a high precision. 

Global methods to model interpretation attempt to provide high-level explanations of global model behavior using a fewer number of rules \cite{lakkaraju2016interpretable, bastani2017interpreting, lakkaraju2019faithful}. In this direction, \cite{lakkaraju2016interpretable} show that a set of independent rules is more interpretable than decision lists (where each rule depends on the negation of the rules before it). MUSE \cite{lakkaraju2019faithful} uses an optimization procedure to optimize an objective function that balances between fidelity and interpretability of the final rule set. The candidate set of rules input to the optimization procedure is derived using association rule mining (Apriori). In contrast, MAGIX uses the combination of LIME and genetic algorithm to construct global rules by optimizing an information theory based fitness measure. EXPLAIN \cite{robnik2018explanation} also extracts conditions similarly using perturbation. \cite{chen2018learning, Kanehira_2019_CVPR} use information-theoretic measures to explain models. Bastani et al. \cite{bastani2017interpreting} and Evans et al. \cite{evans2019s} propose a surrogate model approach where a decision tree is trained on the predictions made by the model. Each path from the root of the tree to a leaf represents a rule and the rule set is an interpretation of model behavior. We compare various existing approaches to MAGIX in Section \ref{results}.

Recent work has demonstrated that post hoc model interpretation methods are vulnerable to adversarial attacks as small perturbations to the input can substantially change the resulting explanations \cite{ribeiro2018semantically, slack2020fooling}. It has also been demonstrated that current approaches lack robustness to distribution shifts i.e., explanations constructed using a given data distribution may not be valid on out of distribution data \cite{ghorbani2019interpretation, lakkaraju2020fool}. We propose a new parameter to evaluate the performance of various existing interpretation approaches in the scenario of distributional shift and propose a method to improve distributional robustness for our approach in Sections \ref{Uncertainty Analysis} and \ref{Ablation Study 3}. 

\section{Approach}\label{approach}
The input to MAGIX is the dataset used to train the model along with model predictions. MAGIX outputs a set of rules that explain model behavior, as a whole, on a global level. A rule $R_k$ consists of a set of conditions (clause) and a predicted class (prediction) (Table \ref{table:sample-rules-MAGIX}). It explains model behavior for the subset of instances that are covered by the rule. An instance is covered by a rule if its feature values are satisfied by the rule clause. The \textit{coverage($R_k$)} of a rule is the fraction of instances that are covered by $R_k$. An instance is \textit{correctly covered} by a rule if the instance is covered by the rule and the class predicted by the rule is the same as the class predicted by the model for the instance. The \textit{precision($R_k$)} of a rule is the fraction of instances in \textit{coverage($R_k$)} that are correctly covered by $R_k$. The Interpretation of a machine learning model is a set of rules that explains model behavior. The \textbf{Fidelity} \cite{guidotti2018survey} of an interpretation is measured by the fraction of instances for which the interpretation correctly predicts model behavior (Section \ref{Simulated User Study}). 

\begin{figure*}
\includegraphics[width=1.0\textwidth]{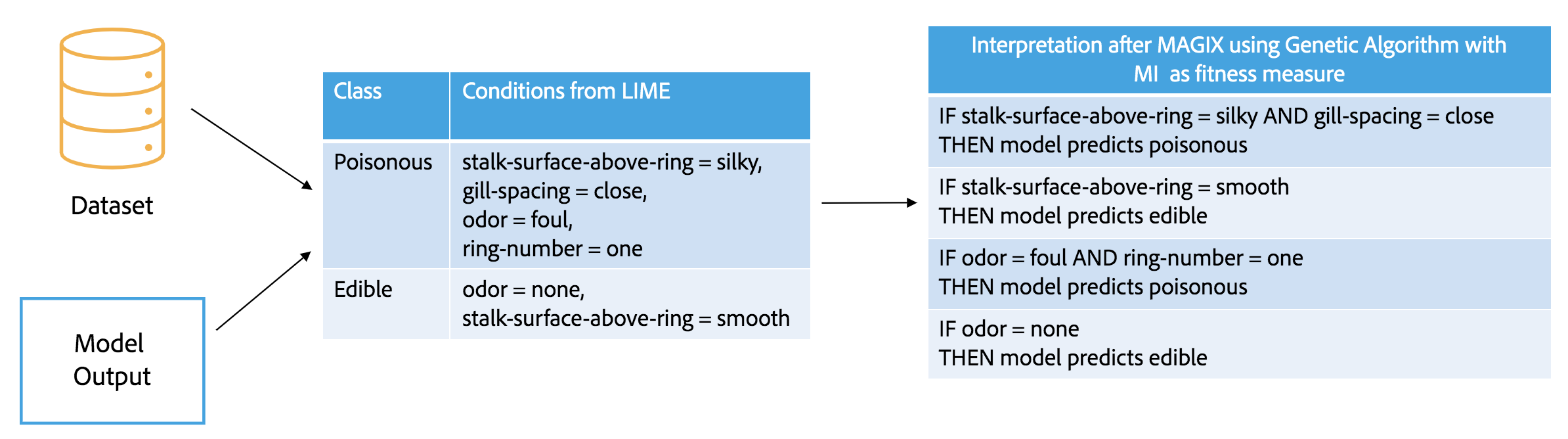}
\centering
\caption{High Level Overview of MAGIX: MAGIX employs LIME \cite{ribeiro2016should} to extract a set of conditions that are important for explaining local model behavior. Different combinations of these conditions are explored using a genetic algorithm guided by a mutual information based fitness measure to derive a set of rules that explain global model behavior.}
\label{figure:magix-high-level}
\end{figure*}

A high level outline of MAGIX is shown in Figure \ref{figure:magix-high-level}. In the condition mining phase, MAGIX employs LIME \cite{ribeiro2016should} to extract a set of conditions that are important for explaining local model behavior. These conditions are then input to a rule construction phase. In this phase, different combinations of these conditions are explored using a genetic algorithm guided by a mutual information based fitness measure to derive a set of rules that explain model behavior.

In the condition mining phase, MAGIX uses LIME \cite{ribeiro2016should} to find conditions that are important to explain the classification decisions at a local level. The output of LIME for an instance is the set of conditions along with associated weights that are important for explaining the classification of the instance. The following process is repeated for each predicted class. An instance is selected at random from the set of instances classified into this class. LIME is run to explain the classification of the selected instance. All conditions output by LIME that have positive weights are added to the set of candidate conditions. This process is repeated until each instance classified into this class is covered by at least one condition. The output of this step is a set of conditions that are locally important for explaining the classification of instances in the dataset. In Section \ref{Ablation-LIME}, we evaluate the effect of using LIME by comparing it to an alternative approach for extracting conditions.

In the rule construction phase, a Genetic Algorithm \cite{whitley1994genetic} is used to evolve global rules from the extracted conditions. The algorithm is run separately for each class. It explores candidate rules i.e combinations of conditions, guided by a fitness function. An individual is encoded as a bit string that represents a rule. For example, if the number of conditions generated for a class $K$ is 10, then each individual of the population for that class is a bit string of length 10. For instance, the individual `1001000000' represents the rule `If $condition_1$ AND $condition_4$ THEN predict Class $K$'. For the numerical variables, MAGIX allows each numerical variable to take on a value from a possible range of values within the candidate rule clause. For the categorical variables in the data, MAGIX allows each categorical variable to take on a value from a possible subset of values within the candidate rule clause. For example, consider a categorical variable `country'. If `$10\leq age < 25$', `$country = US$', and `$country = India$' are conditions for class $K$, then `If $10\leq age < 25$ AND $country \in \{US, India\}$ THEN Predict Class $K$' is a valid candidate rule. This improves the quality of human interpretation by capturing more patterns with fewer rules.

The initial population for the genetic algorithm consists of randomly initialized individuals. The individuals chosen for mating in each generation are selected using a k-way tournament with k = 3. Uniform crossover is used for mating individuals and the old generation is replaced completely by the new generation. All other GA hyper parameters are selected as described in Section \ref{results} determined using grid search over values tabulated in Table \ref{table:hyper-param-GA}. The algorithm is run for a number of generations and the final population (set of rules) is the interpretation of model behavior for a particular class. It is repeated for each predicted class to get the final result. In Section \ref{Ablation-GA}, we evaluate the effect of using a genetic algorithm in the rule construction phase by comparing it to alternative approaches for learning rules.

A rule that explains model behavior with high fidelity has high \textbf{precision} and \textbf{coverage}. Therefore, the genetic algorithm in the rule construction phase needs a fitness function that captures these characteristics. The natural candidate for the fitness function is the $F_1$ score, i.e. the harmonic mean of precision and coverage. However, the $F_1$ score performs poorly for imbalanced datasets as explained in Section \ref{mi} and shown in Section \ref{Ablation-MI}. Therefore, we propose a fitness measure based on mutual information. We compare the effect of using different fitness measures in Section \ref{Ablation-MI}.

\subsection{Mutual Information as Fitness Measure}\label{mi}
The Mutual Information ($MI$) between two variables quantifies the information one variable provides about the other \cite{cover2012elements}. Within the context of a rule, the two variables are the model behavior and the rule prediction. Concretely, we want to answer the question `How much information does a rule provide about the behavior of the model?'. We want to maximize the information that each rule provides us about the model's behavior. For a rule $R_i$ that has class label $y_{R_i}$, we construct a contingency table as shown in Table \ref{table:rule-mutual-information}. It is important to note that `Class' refers to the class predicted by the model, and not the ground truth class label (as explained at the beginning of Section \ref{approach}).

Mutual Information for a rule $R_i$ is calculated from the contingency table as:
 
\begin{equation}\label{formula:MI}
MI = \dfrac{1}{N} \sum_{a, b = 1}^{2}n_{ab}log(\dfrac{n_{ab}\times N}{r_a\times c_b})
\end{equation}

where,\\
N is the sum of all values\\
$r_a$ is the sum of values in row a\\
$c_b$ is the sum of values in column b\\

The fitness measure for a rule is:
\begin{equation}\label{formula:Fitness}
Fitness(R_i) = \left\{ \begin{array}{rl} MI & when\;n_{11} \geq \dfrac{r_1\times c_1}{N} \\
-1\times MI & otherwise \end{array}\right.
\end{equation}

To understand why the fitness correctly captures rules that explain model behavior, consider the example rules (that predict class 0) given in Tables \ref{table:rule-mutual-information-explain-case-1} through \ref{table:rule-mutual-information-explain-case-4}. For this example, the dataset has 2000 instances and the model predicts class 0 for 1600 instances and class 1 for 400 instances.

\subsubsection{Case 1: Rules that do not explain model behavior}
Table \ref{table:rule-mutual-information-explain-case-3} illustrates a rule whose predictions do no better than random at explaining model behavior. To understand this, compare this rule to one that explains model behavior by assigning a single class to a randomly selected subset of data. Such a rule has a class distribution similar to the class distribution of the dataset. For instance, if the model predicts class 0 for 80\% of instances and class 1 for 20\% of instances, then any rule that predicts class 0 for a random subset of data will correctly explain 80\% of covered instances and have a precision of 0.8. However, the rule provides little information about patterns used by the model. This is captured by the $MI$ of the rule, that is equal to 0. In general, for any rule that does no better than random at explaining model behavior, the value of $n_{11}$ is close to the expected value $\dfrac{r_1\times c_1}{N}$, and the $MI$ and subsequently fitness is close to 0. Therefore, such rules are not output by the genetic algorithm. 

It is important to note that the same is not true for most other metrics based on precision and coverage of the rule. For example, the $F1$-score, which is the harmonic mean of precision and coverage, is high when both precision and coverage are high. For example, the rule in Table \ref{table:rule-mutual-information-explain-case-3} has a high $F_1$ score. In general, in datasets with class imbalance, the problem with metrics based on precision and coverage becomes more pronounced. This is shown through an ablation study in Section \ref{Ablation-MI}.  

\subsubsection{Case 2: Rules that explain model behavior}
Table \ref{table:rule-mutual-information-explain-case-1} illustrates a rule whose predictions match those made by the model. It is therefore good at explaining model behavior. In general, a high value of $MI$ for a rule implies that the rule accurately explains model behavior. Such rules have a high fitness value and are output by the genetic algorithm.  

\subsubsection{Case 3: Rules that contradict model behavior}
Table \ref{table:rule-mutual-information-explain-case-4} illustrates a rule whose predictions contradict those made by the model. It is therefore poor at explaining model behavior. However, the rule has a high value of $MI$, since it provides information about the model behavior, albeit in a negative direction. To penalize such rules, we negate the value of $MI$ to compute rule fitness when the value in cell $n_{11}$ is less than the expected value. This is illustrated in Equation \ref{formula:Fitness}. 

\begin{table}
\caption{Contingency table used to compute \textit{Fitness score} of a candidate rule}
\begin{tabular}{|p{.08\textwidth}|p{.16\textwidth}|p{.16\textwidth}|}
\hline
Rule/Class & $y_{R_i}$  & NOT $y_{R_i}$\\ \hline
$R_i$ & $n_{11}$ = Number of instances in cover($R_i$) with class $y_{R_i}$ & $n_{12}$ = Number of instances in cover($R_i$) with class different from $y_{R_i}$\\ \hline
NOT $R_i$ & $n_{21}$ = Number of instances with class $y_{R_i}$ but not covered by $R_i$ & $n_{22}$ = Number of instances not covered by $R_i$ and with class different from $y_{R_i}$\\ \hline
\end{tabular}
\label{table:rule-mutual-information}
\end{table}

\begin{table}
\caption{Rule that explains model behavior\\$MI=0.118$ \space \space$F1=0.545$ \space \space$Fitness=0.118$}
\begin{tabular}{|p{.08\textwidth}|p{.16\textwidth}|p{.16\textwidth}|}
\hline
Rule/Class & Class $0$  & Class $1$\\ \hline
$R_1$ & $n_{11}$ = 600 & $n_{12}$ = 0\\ \hline
NOT $R_1$ & $n_{21}$ = 1000 & $n_{22}$ = 400 \\ \hline
\end{tabular}
\label{table:rule-mutual-information-explain-case-1}
\end{table}

\begin{table}
\caption{Rule that does not explain model behavior\\$MI=0.0$ \space \space$F1=0.615$ \space \space$Fitness=0.0$}
\begin{tabular}{|p{.08\textwidth}|p{.16\textwidth}|p{.16\textwidth}|}
\hline
Rule/Class & Class $0$  & Class $1$\\ \hline
$R_2$ & $n_{11}$ = 800 & $n_{12}$ = 200\\ \hline
NOT $R_2$ & $n_{21}$ = 800 & $n_{22}$ = 200 \\ \hline
\end{tabular}
\label{table:rule-mutual-information-explain-case-3}
\end{table}

\begin{table}
\caption{Rule negatively correlated with model behavior\\$MI=0.118$ \space \space$F1=0.667$ \space \space$Fitness=-0.118$}
\begin{tabular}{|p{.08\textwidth}|p{.16\textwidth}|p{.16\textwidth}|}
\hline
Rule/Class & Class $0$ & Class $1$\\ \hline
$R_3$ & $n_{11}$ = 1000 & $n_{12}$ = 400\\ \hline
NOT $R_3$ & $n_{21}$ = 600 & $n_{22}$ = 0 \\ \hline
\end{tabular}
\label{table:rule-mutual-information-explain-case-4}
\end{table}

\section{Results}\label{results}

To show that MAGIX learns an interpretation that correctly explains model behavior, we use the simulated user study method of comparison described in \cite{ribeiro2018anchors}. In Section \ref{Ablation-LIME}, we evaluate alternative approaches for the condition mining phase of MAGIX. In Section \ref{Ablation-MI}, we evaluate alternative fitness measures that can be used to guide the genetic algorithm. In Section \ref{Ablation-GA}, we evaluate alternative approaches to build rules using locally important conditions. 

In Section \ref{Uncertainty Analysis}, we show how the quality of interpretation extracted by existing approaches falls when we evaluate them on distributions that the model has not seen previously. This indicates that the interpretations lack distributional robustness and rather capture aspects of the data distribution that are not learned by the model. In Section \ref{Ablation Study 3}, we show how augmenting the original dataset with out of distribution samples prior to running MAGIX leads to an interpretation that is more robust to distributional shifts.

For all experiments, each dataset is split into a training (60\%), validation (20\%) and scoring (20\%) set. Models are trained on the training set. Model hyper parameters are determined using grid search \cite{scikit-learn} to optimise accuracy on the validation set. For each numerical dataset, a Random Forest model \cite{scikit-learn} is used. For datasets that have categorical features along with numerical features, a Gradient Boosting model is used \cite{light-gbm}. The interpretation approaches are run on the predictions made by the black box model on the training set. Numerical features are discretized prior to running the interpretation approaches using entropy-based binning \cite{scikit-learn}. For LIME, we use the implementation in \cite{ribeiro2016should}. For implementing the Genetic Algorithm, we use \cite{DEAP_JMLR2012}. 

All interpretation hyper parameters are determined using grid search (Table \ref{table:hyper-param-GA}) to optimise \textit{Set-Score} (Definition \ref{set-score}) on the validation set. The evaluation metrics are computed on the held out scoring set. Table \ref{table:Dataset-Description} lists the datasets used for our experiments \cite{Dua:2017}. These datasets allow us to evaluate the performance of MAGIX as well as existing approaches across a wide range of number of samples, features, classes and data types.

\begin{table}
\centering
\caption{Dataset Descriptions}
\begin{tabular}{|p{0.09\textwidth}|p{0.05\textwidth}|p{0.06\textwidth}|p{0.05\textwidth}|p{0.09\textwidth}|}
    \hline
    Dataset&Rows&Features&Classes&Type \\
    \hline
    NBA & 1340 & 19 & 2 & Numerical \\ 
    \hline
    Wi-Fi & 2000 & 7 & 4 & Numerical \\
    \hline
    Statlog & 58000 & 9 & 7 & Numerical\\
    \hline
    Forest & 54000 & 54 & 6 & Mixed\\
    \hline
    Abalone & 4177 & 8 & 29 & Mixed\\
    \hline
    Character & 6000 & 7 & 10 & Categorical\\
    \hline
    Car & 1728 & 6 & 4 & Categorical\\
    \hline
    Chess & 28056 & 6 & 17 & Categorical\\
    \hline
    Mushroom & 8124 & 21 & 2 & Categorical\\
    \hline
    Tic-Tac-Toe & 958 & 9 & 2 & Categorical\\
    \hline
\end{tabular}
\label{table:Dataset-Description}
\end{table}

\begin{table}[]
\centering
\caption{Grid search parameters for all experiments}
\begin{tabular}{|p{.07\textwidth}|p{.17\textwidth}|p{.17\textwidth}|}
\hline
\textbf{Stage} & \textbf{Hyperparameter name} & \textbf{Range} \\ \hline
\multirow{4}{*}{\shortstack[l]{Genetic\\ Algorithm}}
& Number of Generations & 1000, 1500, 2000, 2500 \\
& Population size & 600, 900, 1200, 1500 \\
& Crossover Probability & 0.2, 0.25, 0.3, 0.35 \\
& Mutation Probability & 0.15, 0.2, 0.25, 0.3 \\ \hline
Decision Tree & Max Depth & 4, 5, 6, 7, 8, 9, 10 \\ \hline
Apriori & Support Threshold &  1, 2, 5, 10 (in \%) \\ \hline
MUSE \cite{lakkaraju2019faithful}  & $\lambda_i$ in Objective function &  Ternary search over range [0, 1000] \\ \hline
\end{tabular}
\label{table:hyper-param-GA}
\end{table}

\subsection{Simulated User Study}\label{Simulated User Study}
We compare the performance of MAGIX to the following:
\begin{enumerate}
    \item Anchors \cite{ribeiro2018anchors}: Anchors generates rules to explain model behavior for a specific instance. To generate a global set of rules, we use the approach suggested in their paper. Anchors is run for each instance in the dataset resulting in a set of rules that explain model behavior in all parts of the state space. 
    \item Apriori \cite{agrawal1996fast}: Association Rule Mining is performed on the predictions of the black box model to construct rules that capture patterns picked up by the model. We use the implementation in \cite{apriori}.
    \item MUSE \cite{lakkaraju2019faithful}: Apriori algorithm is used to find candidate sets of conjunctions of predicates i.e. candidate rules. This same candidate set is assigned to both neighborhood descriptors (ND) as well as decision logic rules (DL). These are input to an optimization search using the procedure and the objective function as defined in their approach \cite{lakkaraju2019faithful} to find a final set of rules that explain global model behavior balancing between fidelity, interpretability and ambiguity. 
    \item Surrogate model Decision Tree based approach (DT) \cite{evans2019s}: A decision tree is trained on the predictions made by the model. Each path from the root of the tree to a leaf is a rule. The interpretation consists of all such rules extracted from the tree.
\end{enumerate}
 
\begin{definition}{}\label{set-score}
Set-Score: The set score of an Interpretation $I$, is the fraction of instances in the dataset $D$ for which $I$ is able to correctly predict model behavior. 
\end{definition}

\textit{Set-Score} is the fraction of instances in the dataset for which the interpretation correctly predicts model behavior. \textit{Set-Score} is a measure to quantify the fidelity of a given interpretation to the black box model. As discussed in Section \ref{mi}, our proposed fitness measure allows us to answer the question `How much information does a rule provide about the behavior of the model?' while evolving the interpretation, thus allowing us to effectively explore and make choices between candidate rules. \textit{Set-Score} allows us to measure the final effect and performance of employing this fitness measure on the scoring set. 

It is computed as follows. For each instance in the scoring set, the interpretation assigns a class that is equal to the class of the highest precision rule in the interpretation that covers this instance. Instances that are not covered by any rule are not assigned a class by the interpretation and therefore contribute negatively towards the Set-Score of the interpretation. The class predicted by the interpretation is then compared to the one assigned by the model for the instance. The percentage of instances where these match is the \textit{Set-Score}. An interpretation that predicts model behavior correctly for all instances in the scoring dataset will have a \textit{Set-Score} of 100. A higher \textit{Set-Score} indicates that the interpretation is able to correctly predict the model behavior for a larger fraction of the scoring dataset. The analogy of this metric to a user study is introduced in \cite{ribeiro2018anchors}. The `simulated' user is provided the set of rules that explain model behavior. The user then has to predict the model behavior on a previously unseen dataset using this set of rules. The algorithm for using the rules to make predictions is predefined. For each instance, it involves selecting the highest precision rule that covers the instance and assigning it the class predicted by that rule.

Table \ref{table:results-simulated-user-study} shows the \textit{Set-Score} of the approaches on different datasets. As we want our interpretation to be human understandable, it is desirable to explain a large part of model behavior with high fidelity/accuracy using a small number of rules. To evaluate and compare the performance of interpretation approaches across different levels of fixed rule set sizes, we use the following method. The rule to be added to the rule set at each step is selected so that the \textit{Set-Score} of the resulting set of rules on the validation set is maximised. This is a variant of the technique described in \cite{ribeiro2016should}. It is applied to the output of each interpretation approach to select a fixed number of rules, thereby allowing us to bring all techniques to an equal footing for comparison. 

\begin{table}
\centering
\caption{Simulated User Study for different sizes of human understandable rule sets}
\begin{tabular}{{|p{0.08\textwidth}|p{0.07\textwidth}|p{0.05\textwidth}|p{0.05\textwidth}|p{0.05\textwidth}|p{0.05\textwidth}|}}
\hline
Dataset & Approach  & \multicolumn{4}{c|}{Set Score} \\ 
\hline
{} & {} & 5 \newline Rules & 10 Rules & 15 Rules & 20 Rules \\ \hline
\multirow{5}{*}{NBA} 
& MAGIX & 72.76 & \textbf{81.72} & \textbf{81.72} & \textbf{81.72} \\
& MUSE & 75.81 & 75.90 & 76.12 & 76.12 \\
& Apriori & \textbf{78.35} & 79.85 & 79.95 & 79.85 \\
& DT & 45.15 & 51.86 & 56.71 & 57.46 \\
& Anchors & 37.78 & 50.00 & 58.52 & 61.48 \\ 
\hline
\multirow{5}{*}{Wi-Fi} 
& MAGIX & \textbf{78.25} & \textbf{92.75} & \textbf{95.25} & \textbf{95.00} \\
& MUSE & 60.50 & 66.00 & 66.20 & 66.80 \\
& Apriori & 78.00 & 91.25 & 92.50 & 93.5 \\
& DT & 44.00 & 44.25 & 45.25 & 69.75 \\
& Anchors & 63.43 & 80.59 & 85.82 & 89.05 \\ 
\hline
\multirow{5}{*}{Statlog} 
& MAGIX & \textbf{95.77} & \textbf{99.51} & \textbf{99.64} & \textbf{99.65} \\
& MUSE & 62.72 & 62.85 & 62.95 & 63.22 \\
& Apriori & 78.77 & 79.14 & 79.56 & 79.77 \\
& DT & 94.15 & 94.15 & 94.15 & 94.15 \\
& Anchors & 53.33 & 60.00 & 73.33 & 73.33 \\ 
\hline
\multirow{5}{*}{Forest} 
& MAGIX & \textbf{70.26} & \textbf{71.76} & \textbf{71.76} & \textbf{71.76} \\
& MUSE & 18.53 & 19.42 & 19.61 & 19.83 \\
& Apriori & 18.50 & 19.40 & 19.61 & 19.74 \\
& DT & 32.05 & 32.88 & 32.99 & 33.09 \\
& Anchors & 22.73 & 31.90 & 38.72 & 43.44 \\ 
\hline
\multirow{5}{*}{Abalone} 
& MAGIX &  \textbf{84.81} & \textbf{89.71} & \textbf{89.95} & \textbf{90.07}\\
& MUSE & 14.47 & 16.57 & 18.77 & 19.47 \\
& Apriori & 32.29 & 33.17 & 33.49 & 33.56 \\
& DT & 23.20 & 30.86 & 42.58 & 49.04 \\
& Anchors & 33.88 & 34.72 & 41.62 & 42.81 \\ 
\hline
\multirow{5}{*}{Character} 
& MAGIX &  26.91 & \textbf{35.02} & \textbf{39.29} & \textbf{41.98}\\
& MUSE & 18.78 & 19.18 & 19.48 & 18.78 \\
& Apriori & \textbf{28.67} & 34.41 & 35.11 & 36.61 \\
& DT & 16.79 & 20.32 & 23.60 & 30.63 \\
& Anchors & 7.41 & 11.08 & 13.02 & 13.96 \\ 
\hline
\multirow{5}{*}{Car} 
& MAGIX &  88.43 & 92.77 & 92.77 & 92.77\\
& MUSE & 84.97 & 86.99 & 86.99 & 86.99 \\
& Apriori & \textbf{95.08} & \textbf{95.48} & \textbf{95.53} & \textbf{95.71} \\
& DT & 61.85 & 71.09 & 74.85 & 77.74 \\
& Anchors & 28.30 & 39.62 & 47.16 & 50.94 \\ 
\hline
\multirow{5}{*}{Chess} 
& MAGIX &  \textbf{67.76} & \textbf{74.96} & \textbf{76.46} & \textbf{76.47}\\
& MUSE & 24.87 & 25.47 & 26.19 & 28.12 \\
& Apriori & 67.16 & 67.37 & 67.41 & 67.82 \\
& DT & 49.41 & 56.95 & 62.86 & 65.11\\
& Anchors & 9.18 & 9.18 & 11.22 & 14.28 \\ 
\hline
\multirow{5}{*}{Mushroom} 
& MAGIX &  \textbf{96.86} & \textbf{98.03} & \textbf{98.03} & 98.03 \\
& MUSE & 75.81 & 76.11 & 76.23 & 76.47 \\
& Apriori & 81.29 & 81.32 & 81.41 & 81.61 \\
& DT & 91.56 & 91.56 & 91.56 & 91.56 \\
& Anchors & 77.95 & 91.93 & 97.26 & \textbf{99.04} \\ 
\hline
\multirow{5}{*}{Tic-Tac-Toe} 
& MAGIX & \textbf{91.14} & \textbf{92.18} & \textbf{92.18} & \textbf{92.18} \\
& MUSE & 80.21 & 89.06 & 89.06 & 89.06 \\
& Apriori & 89.06 & 90.62 & 90.72 & 90.86 \\
& DT & 51.56 & 65.10 & 74.48 & 77.60 \\
& Anchors & 33.58 & 50.75 & 61.19 & 68.65 \\ 
\hline

\end{tabular}
\label{table:results-simulated-user-study}
\end{table}

MAGIX outperforms existing approaches for most datasets and across different levels fixed rule set sizes. Association Rule Mining (Apriori) comes close to MAGIX on some datasets that are simpler i.e., datasets that have a small number of rows, features and classes with numerical data. However, it performs poorly on most complex datasets i.e., datasets having a large number of rows with some features having categorical data. Anchors has a poor value for the \textit{Set-Score} metric because it generates high precision rules in the locality of various instances being explained. Therefore, the rules in the interpretation have high precision but poor coverage and subsequently the interpretation has a low \textit{Set-Score}. Due to reasons similar to Anchors, MUSE also has poor coverage and needs a much larger number of rules than MAGIX to achieve the same coverage level. Decision Tree (DT) has a poor performance for most datasets.

\subsection{Ablation 1: Alternatives to LIME}\label{Ablation-LIME}

To evaluate the importance of the local condition extraction step of MAGIX, we compare it to an approach using frequent condition mining (Apriori) for condition extraction followed by the genetic algorithm with our mutual information based fitness function. With Apriori, conditions that cover a higher fraction of instances (for a particular class) than a certain support threshold constitute the candidate conditions for that class. For this comparison, we have used Apriori with 1\% and 5\% support thresholds.

The results are shown in Table \ref{table:results-ablation-study-LIME}. For most datasets, the interpretation generated by using local condition extraction with LIME achieves a higher value of \textit{Set-Score}.

\begin{table}
\centering
\caption{Ablation Study 1: Alternatives to LIME}
\begin{tabular}{{|p{0.08\textwidth}|p{0.09\textwidth}|p{0.05\textwidth}|p{0.05\textwidth}|p{0.05\textwidth}|p{0.05\textwidth}|}}
\hline
Dataset & Approach  & \multicolumn{4}{c|}{Set Score} \\ 
\hline
{} & {} & 5 \newline Rules & 10 Rules & 15 Rules & 20 Rules \\ \hline
\multirow{3}{*}{NBA} 
& LIME& \textbf{72.76} & \textbf{81.72} & \textbf{81.72} & \textbf{81.72} \\
& Apriori (1\%) & 65.29 & 70.15 & 71.64 & 72.01 \\
& Apriori (5\%) & 67.54 & 70.89 & 71.27 & 71.64 \\
\hline
\multirow{3}{*}{Wi-Fi} 
& LIME& \textbf{78.25} & \textbf{92.75} & \textbf{95.25} & \textbf{95.00} \\
& Apriori (1\%) & 77.75 & 93.00 & 94.00 & 93.75 \\
& Apriori (5\%) & 77.50 & 86.00 & 89.25 & 89.75 \\
\hline
\multirow{3}{*}{Statlog} 
& LIME& \textbf{95.77} & \textbf{99.51} & \textbf{99.64} & \textbf{99.65} \\
& Apriori (1\%) & 86.63 & 87.42 & 87.33 & 87.34 \\
& Apriori (5\%) & 91.54 & 92.92 & 92.96 & 93.14 \\
\hline
\multirow{3}{*}{Forest} 
& LIME & \textbf{70.26} & 71.76 & 71.76 & 71.76 \\
& Apriori (1\%) & 66.98 & 78.37 & 79.99 & 80.63 \\
& Apriori (5\%) & 66.75 & \textbf{78.63} & \textbf{80.04} & \textbf{80.78} \\
\hline
\multirow{3}{*}{Abalone} 
& LIME &  \textbf{84.81} & \textbf{89.71} & \textbf{89.95} & \textbf{90.07}\\
& Apriori (1\%) & 47.73 & 53.59 & 56.57 & 57.41 \\
& Apriori (5\%) & 41.26 & 49.16 & 49.88 & 50.00 \\
\hline
\multirow{3}{*}{Character} 
& LIME &  \textbf{28.91} & \textbf{36.02} & \textbf{39.49} & \textbf{41.98}\\
& Apriori (1\%) & 24.53 & 34.35 & 38.14 & 39.97 \\
& Apriori (5\%) & 27.30 & 35.99 & 39.35 & 41.54 \\
\hline
\multirow{3}{*}{Car} 
& LIME &  88.43 & \textbf{92.77} & \textbf{93.27} & \textbf{93.77}\\
& Apriori (1\%) & \textbf{91.91} & 92.75 & 93.06 & 93.06 \\
& Apriori (5\%) & 91.04 & 91.04 & 91.04 & 91.04 \\
\hline
\multirow{3}{*}{Chess} 
& LIME &  67.76 & \textbf{77.96} & \textbf{78.46} & \textbf{78.62}\\
& Apriori (1\%) & \textbf{70.72} & 76.55 & 77.83 & 78.59 \\
& Apriori (5\%) & 67.42 & 74.03 & 76.82 & 78.28 \\
\hline
\multirow{3}{*}{Mushroom} 
& LIME &  \textbf{96.86} & \textbf{98.03} & \textbf{98.03} & \textbf{98.03} \\
& Apriori (1\%) & 94.40 & 94.40 & 94.40 & 94.40 \\
& Apriori (5\%) & 96.07 & 97.17 & 97.27 & 98.01 \\ \hline
\multirow{3}{*}{Tic-Tac-Toe} 
& LIME & \textbf{91.14} & \textbf{92.18} & \textbf{92.18} & \textbf{92.18} \\
& Apriori (1\%) & 90.67 & 92.12 & 92.13 & 92.17 \\
& Apriori (5\%) & 91.13 & 91.75 & 91.81 & 92.12 \\ \hline
\end{tabular}
\label{table:results-ablation-study-LIME}
\end{table}

\subsection{Ablation 2: Alternatives to the Mutual Information based fitness measure}\label{Ablation-MI}
To evaluate the importance of the mutual information based fitness measure (Section \ref{mi}), we compare it to a fitness function that is the harmonic mean of the precision and coverage of the rule ($F_1$ score). The conditions output by LIME are used as the input to the genetic algorithm. The fitness function of the genetic algorithm for this study is the $F_1$ score of the rule. The results are shown in Table \ref{table:results-ablation-study-MI}. It can be observed that for the more complex datasets, MI performs better than $F_1$ Score.

\begin{table}
\centering
\caption{Ablation Study 2: Alternatives to the Fitness Measure used by the Genetic Algorithm}
\begin{tabular}{{|p{0.08\textwidth}|p{0.07\textwidth}|p{0.05\textwidth}|p{0.05\textwidth}|p{0.05\textwidth}|p{0.05\textwidth}|}}
\hline
Dataset & Approach  & \multicolumn{4}{c|}{Set Score} \\ 
\hline
{} & {} & 5 \newline Rules & 10 Rules & 15 Rules & 20 Rules \\ \hline
\multirow{2}{*}{NBA} 
& MI & 72.76 & \textbf{81.72} & \textbf{81.72} & 81.72 \\
& $F_1$ & \textbf{74.62} & 79.85 & 81.34 & \textbf{82.08} \\
\hline
\multirow{2}{*}{Wi-Fi} 
& MI & 78.25 & 92.75 & \textbf{95.25} & \textbf{95.00} \\
& $F_1$ & \textbf{80.25} & \textbf{93.50} & 94.75 & 94.82 \\
\hline
\multirow{2}{*}{Statlog} 
&  MI & \textbf{95.77} & \textbf{99.51} & \textbf{99.64} & \textbf{99.65} \\
& $F_1$ & 95.16 & 99.11 & 99.37 & 99.46 \\
\hline
\multirow{2}{*}{Forest} 
& MI & \textbf{70.26} & \textbf{71.76} & \textbf{71.76} & \textbf{71.76} \\
& $F_1$ & 46.45 & 46.48 & 46.48 & 46.48 \\
\hline
\multirow{2}{*}{Abalone} 
& MI & 84.81 & \textbf{89.71} & \textbf{89.95} & \textbf{90.07}\\
& $F_1$ & \textbf{88.63} & 89.47 & 89.47 & 89.47 \\
\hline
\multirow{2}{*}{Character} 
& MI &  \textbf{26.91} & \textbf{35.02} & \textbf{39.29} & \textbf{41.98}\\
& $F_1$ & 26.25 & 33.33 & 38.44 & 40.96 \\
\hline
\multirow{2}{*}{Car} 
& MI &  \textbf{88.43} & 92.77 & 92.77 & 92.77\\
& $F_1$ & 86.70 & \textbf{93.35} & \textbf{93.35} & \textbf{93.35} \\
\hline
\multirow{2}{*}{Chess} 
& MI &  67.76 & \textbf{74.96} & \textbf{76.46} & \textbf{76.47}\\
& $F_1$ & \textbf{68.24} & 73.45 & 74.28 & 74.34 \\
\hline
\multirow{2}{*}{Mushroom} 
& MI &  96.86 & 98.03 & 98.03 & 98.03 \\
& $F_1$ & \textbf{98.89} & \textbf{98.89} & \textbf{98.89} & \textbf{98.89} \\
\hline
\multirow{2}{*}{Tic-Tac-Toe} 
& MI & \textbf{91.14} & \textbf{92.18} & \textbf{92.18} & \textbf{92.18} \\
& $F_1$ & 89.58 & 89.58 & 89.58 & 89.58 \\
\hline
\end{tabular}
\label{table:results-ablation-study-MI}
\end{table}

\subsection{Ablation 3: Alternatives to the Genetic Algorithm}\label{Ablation-GA}

To evaluate the importance of using the genetic algorithm (with the mutual information based fitness measure) in the rule building phase, we compare it to the following alternatives: 
\begin{enumerate}
    \item MUSE: The output of LIME is fed as input to the optimization procedure outlined in MUSE. It optimizes an objective function that balances fidelity, ambiguity and interpretability as outlined in their paper \cite{lakkaraju2019faithful}.
    \item Apriori: The output of LIME is fed as input to Apriori for building rules. Rules that have support greater than a fixed threshold (5\%) are output by the approach.  
\end{enumerate}

The results are shown in Table \ref{table:results-ablation-study-GA}. For most datasets, the interpretation generated by using genetic algorithm with our mutual information based fitness measure achieves a higher value of \textit{Set-Score}. Hence, our mutual information based fitness measure along with the genetic algorithm learns a better interpretation than the alternative rule building approaches.

\begin{table}
\centering
\caption{Ablation Study 3: Alternatives to the Genetic Algorithm}
\begin{tabular}{{|p{0.08\textwidth}|p{0.07\textwidth}|p{0.05\textwidth}|p{0.05\textwidth}|p{0.05\textwidth}|p{0.05\textwidth}|}}
\hline
Dataset & Approach  & \multicolumn{4}{c|}{Set Score} \\ 
\hline
{} & {} & 5 \newline Rules & 10 Rules & 15 Rules & 20 Rules \\ \hline
\multirow{3}{*}{NBA} 
& GA & 72.76 & 81.72 & 81.72 & 81.72 \\
& MUSE & 61.94 & 69.40 & 69.77 & 69.77 \\
& Apriori & \textbf{80.97} & \textbf{83.20} & \textbf{83.20} & \textbf{83.20} \\
\hline
\multirow{3}{*}{Wi-Fi} 
& GA & 78.25 & 92.75 & 95.25 & 95.00 \\
& MUSE & 56.25 & 61.50 & 61.50 & 61.50 \\
& Apriori & \textbf{82.75} & \textbf{93.25} & \textbf{95.50} & \textbf{95.50} \\
\hline
\multirow{3}{*}{Statlog} 
&  GA & \textbf{95.77} & \textbf{99.51} & \textbf{99.64} & \textbf{99.65} \\
& MUSE & 78.47 & 78.62 & 78.62 & 78.62 \\
& Apriori & 79.08 & 79.43 & 79.43 & 79.43 \\
\hline
\multirow{3}{*}{Forest} 
& GA & \textbf{70.26} & \textbf{71.76} & \textbf{71.76} & \textbf{71.76} \\
& MUSE & 18.19 & 18.19 & 18.19 & 18.19 \\
& Apriori & 17.80 & 17.80 & 17.80 & 17.80 \\
\hline
\multirow{3}{*}{Abalone} 
& GA &  \textbf{84.81} & \textbf{89.71} & \textbf{89.95} & \textbf{90.07}\\
& MUSE & 67.10 & 67.10 & 67.10 & 67.10 \\
& Apriori & 70.93 & 71.17 & 71.17 & 71.17 \\
\hline
\multirow{3}{*}{Character} 
& GA &  \textbf{26.91} & \textbf{35.02} & \textbf{39.29} & \textbf{41.98}\\
& MUSE & 20.89 & 20.89 & 20.89 & 20.89 \\
& Apriori & 23.81 & 27.24 & 28.66 & 28.71 \\
\hline
\multirow{3}{*}{Car} 
& GA &  88.43 & 92.77 & 92.77 & 92.77\\
& MUSE & 86.41 & 92.77 & 92.77 & 92.77 \\
& Apriori & \textbf{94.79} & \textbf{94.79} & \textbf{94.79} & \textbf{94.79} \\
\hline
\multirow{3}{*}{Chess} 
& GA &  67.76 & \textbf{74.96} & \textbf{76.46} & \textbf{76.47}\\
& MUSE & 17.83 & 17.83 & 17.83 & 17.83 \\
& Apriori & \textbf{67.80} & 67.80 & 67.80 & 67.80 \\
\hline
\multirow{3}{*}{Mushroom} 
& GA &  \textbf{96.86} & \textbf{98.03} & \textbf{98.03} & \textbf{98.03} \\
& MUSE & 75.75 & 75.75 & 75.75 & 75.75 \\
& Apriori & 95.01 & 95.01 & 95.01 & 95.01 \\
\hline
\multirow{3}{*}{Tic-Tac-Toe} 
& GA & \textbf{91.14} & \textbf{92.18} & \textbf{92.18} & \textbf{92.18} \\
& MUSE & 75.00 & 81.77 & 81.77 & 81.77 \\
& Apriori & 91.14 & 91.14 & 91.14 & 91.14 \\
\hline
\end{tabular}
\label{table:results-ablation-study-GA}
\end{table}

\subsection{Uncertainty Analysis}\label{Uncertainty Analysis}
An interpretation that accurately explains model behavior should be able to imitate the model on previously unseen instances. However, recent work has demonstrated existing post hoc model interpretation approaches to lack distributional robustness i.e., explanations constructed using a given data distribution may not be valid on out of distribution data \cite{ghorbani2019interpretation, lakkaraju2020fool}. One major cause of lack of distributional robustness is the existence of multiple explanations for one part model behavior on the original data distribution, each with the same fidelity \cite{lakkaraju2020fool}. This makes it difficult to choose between such overlapping explanations.

In this experiment, we measure how the \textit{Set-Score} of existing approaches falls when they are evaluated on previously unseen data distributions. This indicates that the explanations generated by these approaches might be capturing patterns of the original data distribution rather than those learned by the model. To measure this, we perturb the original data distribution and measure the \textit{Set-Score} on samples drawn from the perturbed distribution. The \textit{Mean Set-Score} and standard deviation across 10 runs for each approach and each method of perturbation are calculated and recorded in Table \ref{table:Uncertainty Analysis} at a fixed rule set size of 20 rules. The data distribution is perturbed using three following methods:
\begin{enumerate}
    \item \textbf{Method 1}: 10 data partitions are generated by randomly sampling rows from the scoring dataset. The size of each partition is 10\% of the original dataset. 
    \item \textbf{Method 2}: In this method, synthetic instances are generated by sampling a value for each feature based on the frequency distribution of the feature values across the scoring dataset. This is repeated to generate 10 partitions, each having size 10\% of the original dataset.
    \item \textbf{Method 3}: In this method, synthetic instances are generated by sampling a value for each feature from a uniform distribution of the values the feature can take in the scoring dataset. For numerical features, the value is sampled uniformly from the range of possible values for that feature. Similarly, for categorical features, the value is sampled uniformly from the list of possible values. This process is repeated to generate 10 partitions, each having size 10\% of the original dataset. 
\end{enumerate}

Table \ref{table:Uncertainty Analysis} shows the results. There are some datasets for which the interpretation approaches correctly explain model behavior even for previously unseen data distributions. This indicates that the interpretation correctly captures patterns learned by the model. However, for other datasets, the \textit{Set-Score} falls significantly as we perturb the original data distribution (Method 2 and Method 3). This indicates that the interpretation is capturing some aspects of the data distribution that are not used by the model for decision making. We discuss a possible solution to this problem in Section \ref{Ablation Study 3}.

\begin{table}
\centering
\caption{Simulated User Study with rule set size fixed at 20 rules on 3 kinds of out of distribution data to study distributional robustness}
\begin{tabular}{{|p{0.065\textwidth}|p{0.065\textwidth}|p{0.09\textwidth}|p{0.09\textwidth}|p{0.09\textwidth}|}}
\hline
Dataset & Approach  & \multicolumn{3}{c|}{Mean Set Score} \\ 
\hline
{} & {} & Method 1 & Method 2 & Method 3 \\ \hline
\multirow{5}{*}{NBA} 
& MAGIX & \textbf{77.99 $\pm$ 2.37} & \textbf{76.94 $\pm$ 1.18} & 71.90 $\pm$ 3.20 \\
& MUSE & 70.00 $\pm$ 3.80 & 72.13 $\pm$ 2.37 & 50.41 $\pm$ 1.80 \\
& Apriori & 74.89 $\pm$ 1.64 & 67.27 $\pm$ 2.32 & \textbf{74.18 $\pm$ 2.40} \\
& DT & 55.37 $\pm$ 1.95 & 35.04 $\pm$ 2.83 & 19.22 $\pm$ 2.31  \\
& Anchors & 58.88 $\pm$ 2.65 & 27.76 $\pm$ 2.93 & 40.87 $\pm$ 2.39 \\ 
\hline
\multirow{5}{*}{Wi-Fi} 
& MAGIX & \textbf{94.60 $\pm$ 0.13} & \textbf{64.28 $\pm$ 1.43} & \textbf{62.82 $\pm$ 2.85}  \\
& MUSE & 64.25 $\pm$ 2.28 & 39.45 $\pm$ 2.27 & 55.67 $\pm$ 2.13  \\
& Apriori & 93.25 $\pm$ 1.24 & 63.85 $\pm$ 2.07 & 61.65 $\pm$ 2.64  \\
& DT & 72.87 $\pm$ 2.31 & 45.15 $\pm$ 2.09 & 46.85 $\pm$ 1.64 \\
& Anchors & 93.06 $\pm$ 1.28 & 45.39 $\pm$ 1.93 & 46.54 $\pm$ 1.53 \\ 
\hline
\multirow{5}{*}{Statlog} 
& MAGIX & \textbf{99.64 $\pm$ 0.06} & 81.69 $\pm$ 0.22 & 32.98 $\pm$ 0.37 \\
& MUSE & 58.62 $\pm$ 0.44 & 41.43 $\pm$ 0.46 & 13.52 $\pm$ 0.51 \\
& Apriori & 78.77 $\pm$ 0.33 & 82.39 $\pm$ 0.33 & 12.04 $\pm$ 0.21 \\
& DT & 93.88 $\pm$ 0.19 & \textbf{94.38 $\pm$ 0.22} & \textbf{46.87 $\pm$ 0.32} \\
& Anchors & 84.00 $\pm$ 7.42 & 42.67 $\pm$ 9.52 & 28.00 $\pm$ 15.43 \\ 
\hline
\multirow{5}{*}{Forest} 
& MAGIX & \textbf{70.50 $\pm$ 0.43} & \textbf{44.65 $\pm$ 0.42} & \textbf{26.60 $\pm$ 0.53} \\
& MUSE & 18.18 $\pm$ 0.31 & 15.78 $\pm$ 0.36 & 10.75 $\pm$ 0.06 \\
& Apriori & 17.91 $\pm$ 0.26 & 16.50 $\pm$ 0.31 & 10.90 $\pm$ 0.10 \\
& DT & 32.86 $\pm$ 0.38 & 19.64 $\pm$ 0.23 & 13.89 $\pm$ 0.34 \\
& Anchors & 43.01 $\pm$ 0.96 & 07.53 $\pm$ 0.63 & 22.13 $\pm$ 0.64 \\ 
\hline
\multirow{5}{*}{Abalone} 
& MAGIX & \textbf{90.13 $\pm$ 0.81} & \textbf{57.67 $\pm$ 1.90} & 10.85 $\pm$ 1.73 \\
& MUSE & 11.63 $\pm$ 0.87 & 4.22 $\pm$ 0.57 & 4.10 $\pm$ 0.66 \\
& Apriori & 26.27 $\pm$ 1.63 & 17.70 $\pm$ 1.45 & 16.11 $\pm$ 0.91 \\
& DT & 36.38 $\pm$ 1.26 & 24.74 $\pm$ 0.74 & \textbf{21.76 $\pm$ 1.33} \\
& Anchors & 41.18 $\pm$ 1.36 & 2.81 $\pm$ 0.61 & 2.65 $\pm$ 0.57 \\ 
\hline
\multirow{5}{*}{\shortstack[l]{Character}} 
& MAGIX & \textbf{41.53 $\pm$ 0.59} & \textbf{31.39 $\pm$ 0.63} & 26.17 $\pm$ 0.44 \\
& MUSE &18.40 $\pm$ 0.37 & 10.37 $\pm$ 0.23 & 14.08 $\pm$ 0.33 \\
& Apriori & 33.75 $\pm$ 0.63 & 22.32 $\pm$ 0.53 & 13.99 $\pm$ 0.35 \\
& DT & 29.31 $\pm$ 0.63 & 17.88 $\pm$ 0.46 & \textbf{29.91 $\pm$ 0.28} \\
& Anchors & 13.88 $\pm$ 0.64 & 2.63 $\pm$ 0.17 & 2.45 $\pm$ 0.22 \\ 
\hline
\multirow{5}{*}{Car} 
& MAGIX & \textbf{94.10 $\pm$ 1.03} & \textbf{93.26 $\pm$ 1.83} & \textbf{94.16 $\pm$ 1.66} \\
& MUSE & 87.19 $\pm$ 1.19 & 87.51 $\pm$ 1.07 & 87.63 $\pm$ 2.03 \\
& Apriori & 92.45 $\pm$ 1.10 & 92.51 $\pm$ 1.20 & 92.60 $\pm$ 1.22 \\
& DT & 83.06 $\pm$ 1.79 & 83.18 $\pm$ 2.42 & 83.17 $\pm$ 2.15 \\
& Anchors & 66.60 $\pm$ 2.68 & 44.72 $\pm$ 9.62 & 41.13 $\pm$ 6.89 \\ 
\hline
\multirow{5}{*}{Chess} 
& MAGIX & \textbf{76.16 $\pm$ 0.37} & \textbf{73.63 $\pm$ 0.56} & \textbf{52.64 $\pm$ 0.66} \\
& MUSE & 48.89 $\pm$ 0.76 & 49.09 $\pm$ 0.76 & 32.22 $\pm$ 0.65 \\
& Apriori & 68.46 $\pm$ 0.46 & 67.70 $\pm$ 0.50 & 50.35 $\pm$ 1.07 \\
& DT & 63.84 $\pm$ 0.35 & 65.19 $\pm$ 0.46 & 58.94 $\pm$ 0.46 \\
& Anchors & 20.71 $\pm$ 4.28 & 10.10 $\pm$ 3.30 & 4.79 $\pm$ 2.14 \\ 
\hline
\multirow{5}{*}{\shortstack[l]{Mush\\room}}
& MAGIX & 97.39 $\pm$ 0.39 & 72.47 $\pm$ 0.10 & 31.05 $\pm$ 1.06 \\
& MUSE & 78.06 $\pm$ 1.55 & 49.57 $\pm$ 1.74 & 32.33 $\pm$ 0.65 \\
& Apriori & 94.77 $\pm$ 0.39 & 52.33 $\pm$ 1.13 & 54.41 $\pm$ 1.09 \\
& DT & 92.79 $\pm$ 0.54 & \textbf{74.07 $\pm$ 0.74} & \textbf{63.83 $\pm$ 0.98} \\
& Anchors & \textbf{99.85 $\pm$ 0.17} & 49.21 $\pm$ 2.25 & 38.41 $\pm$ 1.13 \\ 
\hline
\multirow{5}{*}{\shortstack[l]{Tic\\ Tac\\ Toe}} 
& MAGIX & \textbf{96.46 $\pm$ 1.09} & \textbf{91.46 $\pm$ 1.55} & \textbf{91.82 $\pm$ 1.69} \\
& MUSE & 89.37 $\pm$ 1.63 & 89.06 $\pm$ 2.51 & 87.45 $\pm$ 2.73 \\
& Apriori & 89.53 $\pm$ 1.95 & 87.92 $\pm$ 2.52 & 86.46 $\pm$ 2.72 \\
& DT & 80.83 $\pm$ 2.69 & 71.45 $\pm$ 2.72 & 74.89 $\pm$ 2.71 \\
& Anchors & 79.40 $\pm$ 3.19 & 59.78 $\pm$ 4.43 & 50.37 $\pm$ 5.71 \\ 
\hline
\end{tabular}
\label{table:Uncertainty Analysis}
\end{table}

\begin{table}
\caption{Ablation Study 4: Simulated User Study with rule set size fixed at 20 rules on 3 kinds of out of distribution data after augmenting the original dataset}
\begin{tabular}{{|p{0.065\textwidth}|p{0.065\textwidth}|p{0.09\textwidth}|p{0.09\textwidth}|p{0.09\textwidth}|}}
\hline
\shortstack[l]{Dataset} & Approach  & \multicolumn{3}{c|}{Mean Set Score} \\ 
\hline
{} & & Method 1 & Method 2 & Method 3 \\ 
\hline

\multirow{5}{*}{NBA}
& MAGIX & \textbf{74.22$\pm$2.48} & \textbf{82.61$\pm$1.47} & \textbf{89.40$\pm$0.87} \\
& MUSE  & 64.66 $\pm$ 2.18 & 70.37 $\pm$ 2.15 & 34.85 $\pm$ 2.09 \\
& Apriori  & 72.46 $\pm$ 2.58 & 78.88 $\pm$ 2.40 & 69.70 $\pm$ 3.06 \\
& DT  & 57.16 $\pm$ 1.98 & 44.55 $\pm$ 3.29 & 28.06 $\pm$ 2.62 \\
& Anchors   & 63.64 $\pm$ 2.89 & 49.51 $\pm$ 3.41 & 50.34 $\pm$ 3.80 \\
\hline

\multirow{5}{*}{Wi-Fi}
& MAGIX & 94.32 $\pm$ 1.42 & 62.92 $\pm$ 2.10 & 61.80 $\pm$ 2.05 \\
& MUSE & 94.15 $\pm$ 1.20 & 70.27 $\pm$ 2.25 & 68.10 $\pm$ 0.85 \\
& Apriori & 95.70 $\pm$ 0.96 & 65.37 $\pm$ 2.57 & 61.55 $\pm$ 2.27 \\
& DT & \textbf{95.95$\pm$0.70} & \textbf{72.97$\pm$2.94} & \textbf{72.37$\pm$1.88} \\
& Anchors & 81.86 $\pm$ 1.54 & 56.64 $\pm$ 1.55 & 57.04 $\pm$ 1.85 \\
\hline

\multirow{5}{*}{Statlog}
& MAGIX & \textbf{97.43$\pm$0.19} & 85.36 $\pm$ 0.26 & 19.27 $\pm$ 0.30 \\
& MUSE & 77.45 $\pm$ 0.37 & 81.61 $\pm$ 0.39 & 7.07 $\pm$ 0.22 \\
& Apriori & 78.61 $\pm$ 0.41 & 84.39 $\pm$ 0.36 & 15.69 $\pm$ 0.21 \\
& DT & 93.70 $\pm$ 0.13 & \textbf{94.54$\pm$0.21} & 50.85 $\pm$ 0.57 \\ 
& Anchors & 81.88 $\pm$ 5.48 & 77.73 $\pm$ 6.30 & \textbf{77.54$\pm$3.09} \\
\hline

\multirow{5}{*}{Forest}
& MAGIX & \textbf{77.45$\pm$0.32} & \textbf{46.84$\pm$0.63} & \textbf{41.69$\pm$0.47} \\
& MUSE  & 18.16 $\pm$ 0.27 & 16.55 $\pm$ 0.31 & 10.14 $\pm$ 0.05 \\
& Apriori  & 20.05 $\pm$ 0.36 & 19.44 $\pm$ 0.32 & 19.16 $\pm$ 0.39 \\
& DT  & 30.86 $\pm$ 0.48 & 17.29 $\pm$ 0.33 & 6.40 $\pm$ 0.22 \\
& Anchors   & 36.14 $\pm$ 0.40 & 16.27 $\pm$ 0.31 & 35.47 $\pm$ 0.71 \\
\hline

\multirow{5}{*}{Abalone}
& MAGIX & \textbf{78.98$\pm$1.30} & \textbf{35.95$\pm$1.25} & 13.83 $\pm$ 1.33 \\
& MUSE & 9.47 $\pm$ 0.94 & 11.29 $\pm$ 1.19 & 5.67 $\pm$ 0.85 \\
& Apriori & 30.13 $\pm$ 2.35 & 28.06 $\pm$ 1.12 & 16.37 $\pm$ 1.67 \\
& DT & 13.22 $\pm$ 1.22 & 11.30 $\pm$ 1.12 & 9.07 $\pm$ 0.66 \\
& Anchors & 54.28 $\pm$ 18.95 & 32.85 $\pm$ 15.71 & \textbf{31.43$\pm$17.84} \\
\hline

\multirow{5}{*}{Character}
& MAGIX & \textbf{40.84$\pm$0.44} & 30.02 $\pm$ 0.66 & 28.87 $\pm$ 0.64 \\
& MUSE & 18.64 $\pm$ 0.53 & 14.59 $\pm$ 0.66 & 14.62 $\pm$ 0.43 \\ 
& Apriori & 24.25 $\pm$ 0.59 & 24.46 $\pm$ 0.45 & 23.82 $\pm$ 0.63 \\
& DT & 26.49 $\pm$ 0.41 & \textbf{38.69$\pm$0.59} & \textbf{61.72$\pm$0.43} \\
& Anchors & 7.50 $\pm$ 0.35 & 3.27 $\pm$ 0.26 & 2.72 $\pm$ 0.23 \\
\hline

\multirow{5}{*}{Car}
& MAGIX & \textbf{99.69$\pm$0.16} & \textbf{98.60$\pm$0.27} & \textbf{99.69$\pm$0.14} \\
& MUSE & 90.29 $\pm$ 1.07 & 91.82 $\pm$ 0.84 & 90.11 $\pm$ 1.85 \\ 
& Apriori & 94.07 $\pm$ 1.48 & 94.13 $\pm$ 1.39 & 93.41 $\pm$ 1.18 \\
& DT & 77.11 $\pm$ 1.64 & 77.08 $\pm$ 2.28 & 76.85 $\pm$ 2.01 \\ 
& Anchors & 81.88 $\pm$ 5.48 & 77.73 $\pm$ 6.48 & 77.54 $\pm$ 3.09 \\
\hline

\multirow{5}{*}{Chess}
& Magix & \textbf{72.52$\pm$0.36} & 69.52 $\pm$ 0.39 & 47.79 $\pm$ 0.64 \\
& MUSE & 45.96 $\pm$ 0.68 & 50.34 $\pm$ 0.55 & 37.21 $\pm$ 0.52 \\ 
& Apriori & 69.67 $\pm$ 0.62 & \textbf{69.86$\pm$0.62} & \textbf{57.38$\pm$0.38} \\
& DT & 60.11 $\pm$ 0.73 & 57.83 $\pm$ 0.57 & 54.86 $\pm$ 0.81 \\
& Anchors & 21.94 $\pm$ 4.48 & 10.51 $\pm$ 2.74 & 12.65 $\pm$ 2.15 \\
\hline

\multirow{5}{*}{\shortstack[l]{Mush\\room}}
& MAGIX & \textbf{99.69$\pm$0.16} & \textbf{98.60$\pm$0.27} & \textbf{99.69$\pm$0.14} \\
& MUSE  & 79.11 $\pm$ 0.77 & 50.91 $\pm$ 0.88 & 40.46 $\pm$ 1.10 \\
& Apriori & 98.06 $\pm$ 0.26 & 83.93 $\pm$ 0.60 & 61.53 $\pm$ 1.42 \\
& DT & 92.96 $\pm$ 0.70 & 74.93 $\pm$ 0.89 & 64.24 $\pm$ 1.07 \\
& Anchors & 91.13 $\pm$ 0.66 & 81.47 $\pm$ 0.73 & 73.23 $\pm$ 0.84 \\
\hline

\multirow{5}{*}{\shortstack[l]{Tic\\Tac\\Toe}}
& Magix & \textbf{91.87$\pm$2.73} & 89.42 $\pm$ 1.94 & \textbf{91.40$\pm$2.17} \\
& MUSE & 89.42 $\pm$ 1.99 & 86.40 $\pm$ 2.81 & 82.13 $\pm$ 2.52 \\
& Apriori & 90.41 $\pm$ 1.16 & \textbf{91.40$\pm$2.35} & 91.30 $\pm$ 1.66 \\
& DT & 73.38 $\pm$ 3.69 & 76.47 $\pm$ 3.30 & 78.85 $\pm$ 2.48 \\
& Anchors & 69.70 $\pm$ 4.43 & 61.72 $\pm$ 2.45 & 48.51 $\pm$ 3.30 \\
\hline

\end{tabular}
\label{table:Abalation-Study-3}
\end{table}

\subsection{Ablation 4: Improving distributional robustness}\label{Ablation Study 3}
To improve the robustness of the interpretation generated by various approaches, the training dataset is augmented with the data generated by \textbf{Method 2} and \textbf{Method 3} prior to running the interpretation approaches. It is important to note that the black box model is not retrained here. The model stays the same as before. Only the interpretation approaches are ran again augmented with this synthetic data along with the model's predictions on this synthetic data. The same Uncertainty Analysis as outlined in Section \ref{Uncertainty Analysis} is now performed on the interpretation learned after augmentation of the training dataset.

Table \ref{table:Abalation-Study-3} shows the results of Uncertainty Analysis performed after this augmentation. For datasets where the interpretation is poor at imitating model behavior for previously unseen distributions, augmenting the training dataset prior to running interpretation approaches leads to an interpretation that is better at predicting model behavior for new previously unseen distributions. MAGIX achieves much higher values of \textit{Set-Score} on NBA, Forest, Car and Mushroom datasets after augmentation. MUSE sees a gain in performance only on the Wi-Fi dataset. Anchors sees a gain in performance in NBA, Statlog, Abalone, Car and Mashroom datasets after augmentation but still remains poor in comparison to all other approaches. A key reason for the lack of distributional robustness is the existence of multiple explanations for one part of the model behavior on the original data distribution, each having the same fidelity \cite{lakkaraju2020fool}. We posit that augmentation of the training dataset allows MAGIX to make more robust choices between such overlapping explanations.

\subsection{Case Study: MAGIX Insights Report}\label{case-study-magix-insights-report}
MAGIX has been deployed in a digital marketing platform (Section \ref{introduction}). MAGIX generates interpretation reports for over 100 organisations across 805 personalization campaigns. For instance, a financial product organisation uses the digital marketing platform to personalize experiences on its website. It runs two personalization campaigns, the first serves 9.2 million users and the second serves 2.7 million users. The user profile that the model trains on has close to 3300 attributes. At this scale, it is not possible to understand model behavior by analysing local explanations for a few users. The MAGIX insights report, generated daily, has an average of close to 300 rules that explain model behavior with an average \textit{Set-Score} of 71\%. The organisation uses the report in two ways. First, marketers from the organisation scan the rules to ensure that the model is not using patterns that contradict either domain experts or country laws. If any rule in the report violates either of these conditions, then the model is switched off for further offline analysis. Second, marketers from the organisation use the rules as user segments for further targeting. Each rule represents a user segment that the model learned for serving content. A subset of these segments are saved and then used by the organisation for targeting on other platforms. Overall, MAGIX (through the Insights Report) has considerably improved the adoption of the personalization offerings of the digital marketing platform.

\section{Conclusion}\label{conclusion}
We have presented an approach that evolves rules using an information-theoretic fitness measure to produce high level explanations that explain model behavior globally. It uses LIME to extract locally important conditions followed by a genetic algorithm that optimizes an information theory based fitness measure to explore combinations of these conditions and construct global rules. MAGIX handles both numerical and categorical variables separately. Allowing categorical variables to take on a value from a possible subset of values, just like numerical variables are allowed to take on a value from a possible range of values, improves the quality of human interpretation by capturing more patterns with fewer rules. Our approach outperforms existing approaches on a variety of publicly available datasets that vary widely across number of rows, features, classes and data types. Further, we have introduced a new parameter to evaluate the distributional robustness of an interpretation of model behavior. It is a measure of how well the interpretation is at predicting model behavior for previously unseen data distributions. We show how existing approaches for interpreting models globally often capture patterns of the training data distribution and not those actually learned by the model. Finally, we show how the quality of the interpretation and its distributional robustness can be improved by adding synthetic out of distribution samples to the dataset used to learn the interpretation.

\bibliographystyle{ACM-Reference-Format}
\bibliography{sample-base}


\end{document}